\documentclass[journal]{IEEEtran}

\usepackage{cite}
\usepackage{amsmath,amssymb,amsfonts}
\usepackage{algorithmic}
\usepackage{graphicx}
\usepackage{textcomp}

\graphicspath{{figures}}

\usepackage[T1]{fontenc} 
\usepackage{booktabs}
\usepackage{orcidlink}
\usepackage{multirow}

\newcommand{\age}{\mathrm{a}}
\newcommand{\hatage}{\widehat{\mathrm{a}}}

\begin{document}

\title{Learning Developmental Age from 3D Infant Kinetics Using Adaptive Graph Neural Networks}

\author{
    Daniel Holmberg\orcidlink{0000-0001-5020-7438}, Manu Airaksinen, Viviana Marchi, Andrea Guzzetta,\\ Anna Kivi, Leena Haataja, Sampsa Vanhatalo, and Teemu Roos
    \thanks{Manuscript edited \today. This work was supported by the Research Council of Finland under the ICT 2023: Frontier AI Technologies program (Grant No. 345635). The results presented were produced using computing resources provided by CSC – IT Center for Science. Clinical data collection, 3D video systems, and pose estimation services were funded by grants from the Research Council of Finland (Grant No. 335788), the Finnish Pediatric Foundation (Lastentautien tutkimussäätiö), the Sigrid Juselius Foundation, HUS Children's Hospital research funds, and the Arvo and Lea Ylppö Foundation.}
    \thanks{D. Holmberg and T. Roos are with the Department of Computer Science, University of Helsinki, Helsinki, Finland (email: daniel.holmberg@helsinki.fi; teemu.roos@helsinki.fi).}
    \thanks{M. Airiksainen and S. Vanhatalo are with the Department of Clinical Neurophysiology, Helsinki University Hospital, Helsinki, Finland (email: manu.airaksinen@hus.fi; sampsa.vanhatalo@hus.fi).}
    \thanks{M. Viviana is with the Department of Developmental Neuroscience, IRCCS Fondazione Stella Maris, Pisa, Italy (email: viviana.marchi@fsm.unipi.it). A. Guzzetta is with the Department of Developmental Neuroscience, IRCCS Fondazione Stella Maris, Pisa, Italy, and the Department of Clinical and Experimental Medicine, University of Pisa, Pisa, Italy (email: andrea.guzzetta@fsm.unipi.it).}
    \thanks{A. Kivi and L. Haataja are with the Department of Pediatric Neurology, Helsinki University Hospital and University of Helsinki, Helsinki, Finland (email: anna.kivi@hus.fi; leena.haataja@hus.fi).}
}

\maketitle

\begin{abstract}
Reliable methods for the neurodevelopmental assessment of infants are essential for early detection of problems that may need prompt interventions. Spontaneous motor activity, or `kinetics', is shown to provide a powerful surrogate measure of upcoming neurodevelopment. However, its assessment is by and large qualitative and subjective, focusing on visually identified, age-specific gestures. In this work, we introduce Kinetic Age (KA), a novel data-driven metric that quantifies neurodevelopmental maturity by predicting an infant's age based on their movement patterns. KA offers an interpretable and generalizable proxy for motor development. Our method leverages 3D video recordings of infants, processed with pose estimation to extract spatio-temporal series of anatomical landmarks, which are released as a new openly available dataset. These data are modeled using adaptive graph convolutional networks, able to capture the spatio-temporal dependencies in infant movements. We also show that our data-driven approach achieves improvement over traditional machine learning baselines based on manually engineered features.
\end{abstract}

\begin{IEEEkeywords}
Motion Analysis, Neurodevelopment, Deep Learning, Graph Representation Learning
\end{IEEEkeywords}

\section{Introduction}

\IEEEPARstart{E}{arly} identification of neurodevelopmental problems is essential for supporting lifelong neurocognitive performance. Providing neurodevelopmental follow-up is a challenge on those over 10 percent of infants who are identified to be at developmental risk based on their pre- or postnatal medical adversities~\cite{rosenberg2008prevalence} such as prematurity, birth asphyxia, stroke or metabolic disorders. Addressing these risks early through timely interventions can significantly reduce their long-term impacts on both individuals and society~\cite{novak2017early}.

Assessment of infants' spontaneous movements may provide a sensitive approach to early prediction of the more severe neurodevelopmental problems, such as cerebral palsy (CP).  During the first months of life, the most widely used paradigm is General Movements Assessment (GMA), where trained experts visually identify specific age-varying patterns (gestalts) in the infants' spontaneous movements~\cite{einspieler1997gma}. In the current clinical practise, the GMA paradigm is typically used to predict later development of cerebral palsy by observing infants' expression of discrete gestures called \emph{fidgety movements} at about 3 months of age~\cite{herskind2015effectivegma}.

Since the GMA paradigm relies on visual inspection by well trained experts, several algorithmic approaches have been developed to emulate the recognition of fidgety movements using video recordings or movement sensors~\cite{adde2009video,stahl2012video,rahmati2014video,stoen2017video}. While they are reported to perform reasonably well, their generalization to external datasets remains to be tested. Moreover, their rationale is significantly challenged by the very low signal-to-noise ratio that comes from the strong class imbalance between the scarce training targets (fidgety movements) and the substantial variability in other video content. Therefore, recent studies have explored a variety of machine learning methods using 2D pose estimated data to perform classifications related to abnormality rather than fidgety movements~\cite{chambers2020bayes,mccay2020binaryab,sakkos2021binaryab,mccay2021pose,hashimoto2022automated,zhang2022cerebral,ni2023semi,turner2023classification}.

\begin{figure*}[h]
    \centering
    \includegraphics[width=\textwidth]{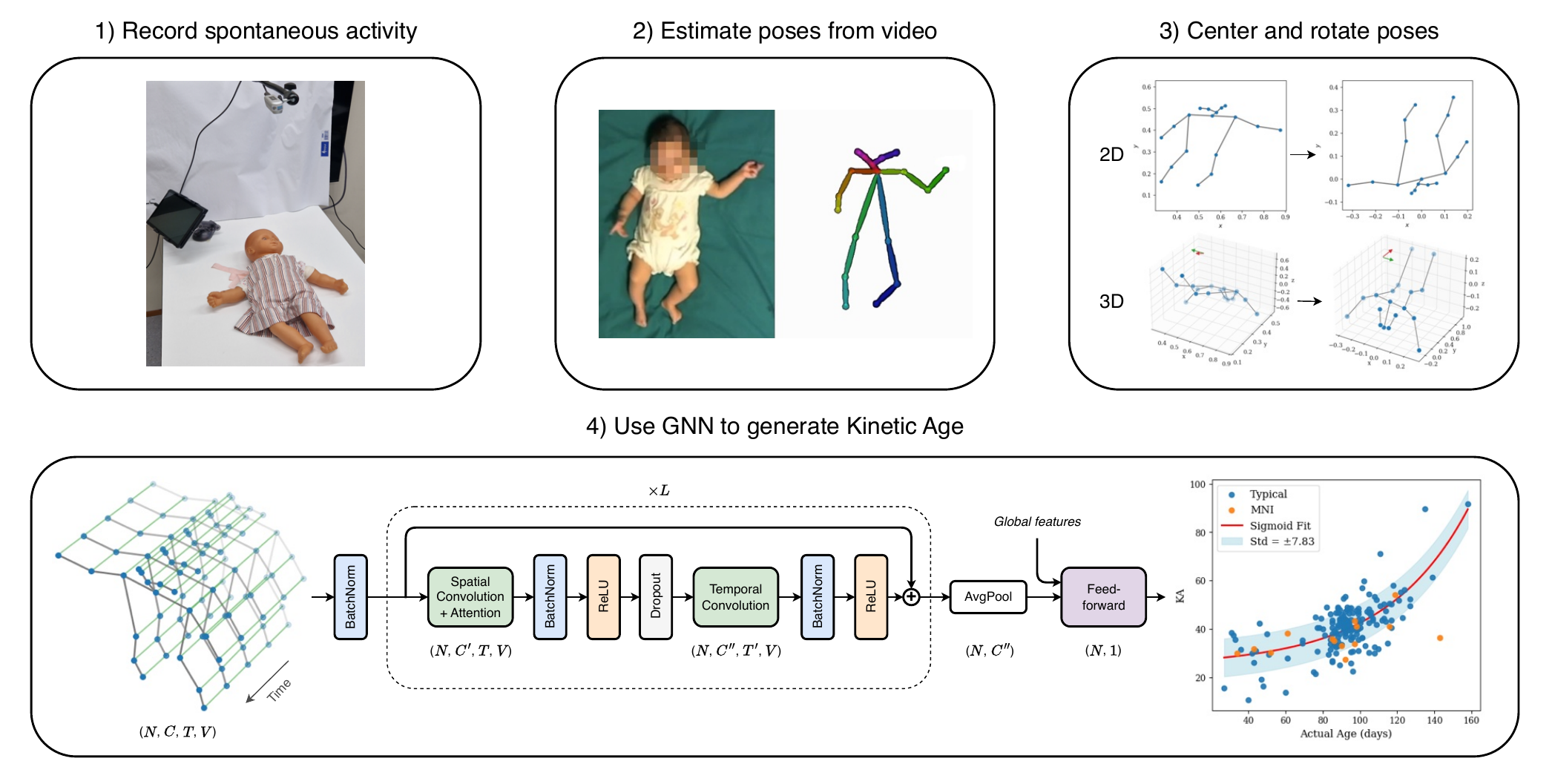}
    \caption{Overview of the study pipeline. The pipeline begins with data collection, where 3D video recordings of spontaneous infant movements are captured using a standardized setup. Pose estimation is applied to generate skeletal videos, extracting spatio-temporal time series of 18 anatomical landmarks. The data is preprocessed through centering and rotation normalization. Finally, Kinetic Age (KA) is predicted using an adaptive graph neural network (GNN) that models spatio-temporal dependencies in the movement patterns.}
    \label{fig:schema}
\end{figure*}

In this work, we propose an alternative approach: instead of emulating experts' evaluation of the GMA assessment or the clinical diagnoses, we choose to target age prediction. It is motivated by the ubiquitous pediatric practice that an infant's age is the key benchmark for evaluating a child's development; that is, determining whether a child is developing typically for the known chronological agently, an age prediction from the spontaneous movements, referred to as kinetic age (KA), in the typically developing infants should provide a maximally generalizable, explainable, and transparent proxy of the maturity in the infants' motor abilities~\cite{stevenson2020automated}. This rationale is conceptualized as functional brain age, which can be computed with the machine learning-based models using physiological or imaging measurements~\cite{stevenson2020automated, airaksinen2023charting, iyer2024growth}.

To this end, we compile a novel dataset from two research centers, consisting of pose-estimated data representing 18 key anatomical landmarks extracted from 3D video recordings. This dataset, which we make openly available, is modeled using Adaptive Graph Convolutional Networks (AAGCNs)~\cite{shi2020aagcn} to capture the spatio-temporal dynamics of infant movements. The model offers notable strengths: it is tailored for skeletal data and includes learnable adjacency matrices and attention mechanisms for each input sample. Through our experiments, we demonstrate that: 1) the data-driven approach offers performance improvements over ML models using previously published hand-crafted movement features, 2) pose estimation obtained in 3D provides a more accurate KA estimate compared to the more broadly studied 2D data, and 3) KA and its derivative KA-gap manage to differentiate between typically and atypically developing infants, offering potential for early clinical diagnosis of neurodevelopmental impairments.

\section{Dataset}

\subsection{Infant Recordings}

The data used in this study were collated from larger research trials conducted in Pisa and Helsinki. Ethical approval for the Italian cohort was provided by the Tuscany Paediatrics Ethics Committee (nr.~187/2020), and for the Finnish cohort, the study protocol was approved by the Institutional Research Review Board at Helsinki University Hospital Medical Imaging Centre. In both countries, written informed consent was obtained from the parents or legal guardians in accordance with the Declaration of Helsinki.

We used 3D video cameras with a tablet-based operating software (Neuro Event Labs Ltd, Tampere, Finland) to record spontaneous movements of infants. The overall setup followed the established protocol by the GMs Trust~\cite{einspieler2005prechtl}. During these sessions, infants were placed in a supine position, and efforts were made to ensure that recordings were captured during periods of active wakefulness while minimizing environmental interferences. Skeletal videos were produced using a computational pose estimation model, and time series of (\textit{x}, \textit{y}, \textit{z})-coordinates from 18 anatomical landmarks were extracted~\cite{marchi2019automatedpose}. The datasets from the two separate institutes could be combined into one larger cohort because they were based on identical video recording devices and the same pose estimation pipeline.

\subsection{Subject Selection and Video Segmentation}

The study cohort included a total of 173 infants from two sub-cohorts: 159 infants with neurodevelopment that was typical or at most  minor neurological impairments, and 14 infants with major neurological impairments (MNI), such as severe CP or cognitive delay, genetic syndrome, or a marked asymmetry. The latter cohort was identified \textit{a priori} by the treating clinicians and it was only used for the last proof of concept pilot. The corrected postnatal ages were comparable between the sub-cohorts: typically developing infants (range-158 days; mean 91.5~$\pm$~21 SD days) and infants with MNI: (range 34--143 days; mean 86.6~$\pm$~30.4 SD days).

\begin{figure*}[t]
    \centering
    \includegraphics[width=\textwidth]{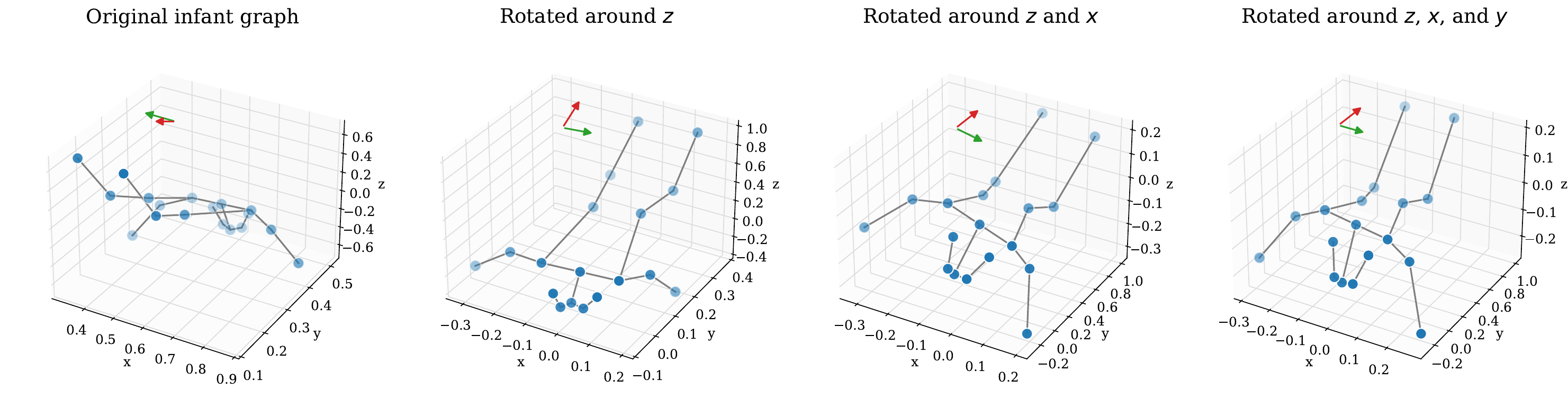}
    \caption{Preprocessing strategy for infant skeletal graphs. The panels shows the average original graph of one recorded segment at each rotation step, with the spine vector indicated in red and the backline vector (average of hipline and shoulder line) in green.}
    \label{fig:preprocessing}
\end{figure*}

Infants within this age range can exhibit a wide variety of behaviors and responses during recordings, including periods of crying, fussiness, or other forms of distress, as well as moments of inactivity or sleep. While these behaviors are natural, they do not provide useful data for analyzing motor kinetics related to developmental milestones. Segments of interest were extracted from the recordings based on annotations provided by pediatricians. This process ensures that the data used for analysis represents appropriate examples of infant movements. The 173 infant time series were divided into a total of 1843 video segments, with 1692 segments from typically developing infants and 151 from the infants with MNI. To standardize the input shapes for spatio-temporal neural networks, the video segments were adjusted to a length of 600 time steps, equivalent to 20 seconds. Segments shorter than 600 time steps were repeated if necessary to meet the length requirement. Baseline machine learning techniques explored in this work are based directly on data from the annotated segments.

\subsection{Rotation Strategy}

To standardize skeletal data across video samples and to minimize variability due to different filming angles, we employ a systematic rotation strategy. An average infant posture before and after these rotations, along with the coordinate system, is depicted in Figure~\ref{fig:preprocessing}. First, we normalize the joint positions by subtracting the neck joint's coordinates from all other joints, establishing a consistent reference frame for all samples.

To ensure comparable orientations across all infants and the video frames, we then performed a sequence of three rotational transformations around the principal axes:  The first rotation aligns the average spine vector, defined from the neck to the center of the hip, with the $y$-axis by rotating around the $z$-axis. In the video image in Figure~\ref{fig:schema}, this would correspond to rotating all infants to the same front facing orientation. Subsequently, we apply a rotation around the $x$-axis to make the average spine parallel to the $y$-axis (head/bottom balance). Finally, to ensure that the spine  (calculated as the average of the hip line and shoulder line) is consistently oriented, we rotate the skeleton around the $y$-axis, aligning the spine with the $x$-axis (left/right balance).

Our method was also used  to process 2D infant poses where only $(x, y)$-coordinates are available. In this case the skeletons are rotated only around the $z$-axis such that the spine vector, defined from the neck to the center of the hip is aligned with $y$-axis. The two subsequent rotations applied on 3D data is not possible for 2D poses where depth (i.e. height from the table) information is missing. Models trained with only $(x, y)$-coordinates are denoted with 2D in the model name, in all other cases models are trained using 3D data.

\subsection{Graph Construction}

\begin{figure*}[ht]
    \centering
    \includegraphics[width=\textwidth]{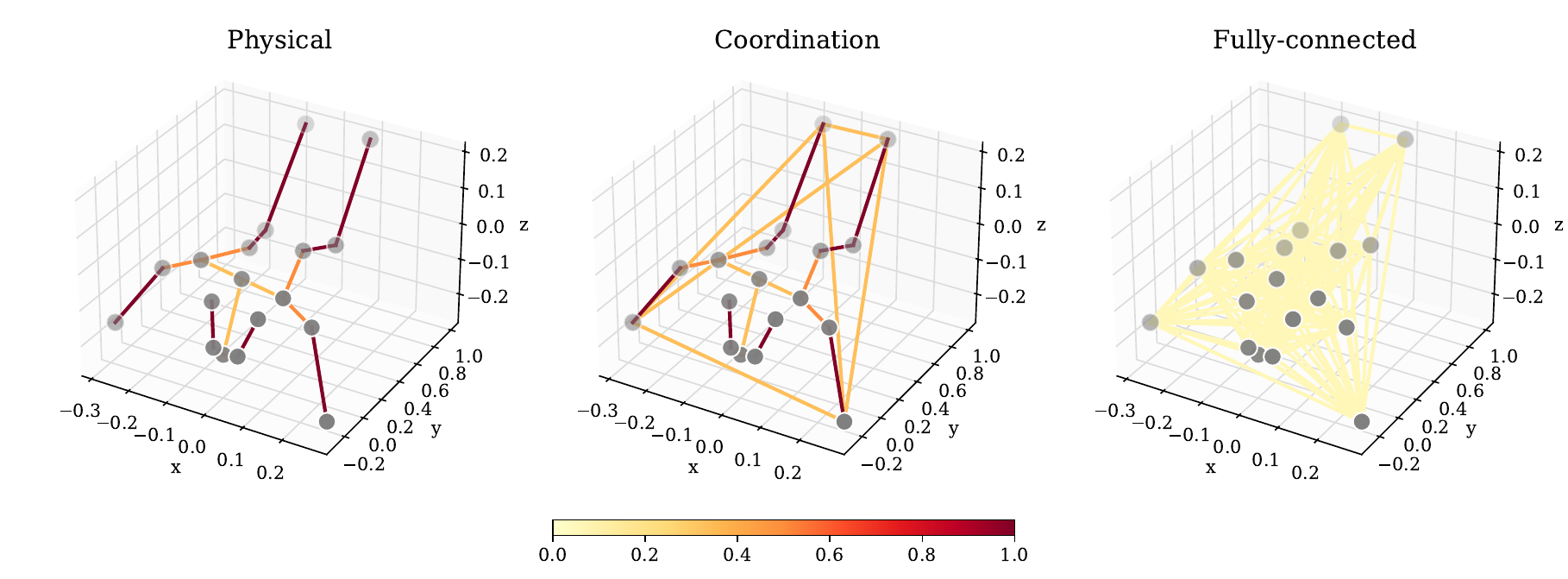}
    \caption{Three different graph initializations for the AAGCN model. The edge weights are colored according to the normalized version of the centrifugal group in the graph partitioning strategy.}
    \label{fig:init}
\end{figure*}

In our model, the human skeleton is represented as a graph, $\mathcal{G} = (\mathcal{V}, \mathcal{E})$, where $\mathcal{V}$ denotes the set of vertices corresponding to human joints, and $\mathcal{E}$ represents the set of edges defining the physical connections between these joints. We consider 18 nodes representing the primary joints extracted through pose estimation, as shown by the blue markers in Figure~\ref{fig:preprocessing}. The natural skeletal structure is modeled using 17 directed edges (gray lines) to reflect the inherent connectivity of the human body. This \emph{physical connectivity initialization} is visualized in the first plot of the first row in Figure~\ref{fig:init}.

To capture coordinated movements often seen in normal infant development, we introduce 6 bidirectional edges between the hands and feet. This is motivated by the observation that normal development requires coordination between limbs: the upper limbs and lower limbs need to work together, such as when hands touch or cross the midline. Additionally, coordination between upper and lower limbs is important for full-body movements like crawling and walking. These \emph{coordination edges} are designed to capture such interactions, as seen in the second plot of Figure~\ref{fig:init}.

As a baseline, we also construct a \emph{fully-connected graph} where every joint is connected to all other joints. This configuration, shown in the third plot of Figure~\ref{fig:init}, removes any inductive bias about the physical structure of the human body.

Finally, to incorporate temporal dynamics, we connect each joint in a given frame to the same joint in the subsequent frame. This spatio-temporal graph structure allows for simultaneous modeling of both spatial relationships among body parts and their temporal evolution throughout the recorded sequences.

\section{Method}
\label{sec:method}

\subsection{Problem Formulation}

The task of predicting an infant's age is formulated as a sequence-to-label problem, where the goal is to infer the infant's corrected age based on their movement patterns. Each infant recording is provided as a multidimensional time series of joint coordinates, which we denote as a tensor $\mathbf{X} \in \mathbb{R}^{C \times T \times V}$, where $C \in \{2, 3\}$ corresponds to the number of spatial coordinates considered, $T=600$ represents the number of temporal frames in the sequence, and $V=18$ is the selected number of joints in the infant graph.

\subsection{Graph Partitioning}

Spatio-temporal Graph Convolutional Networks (ST-GCN)~\cite{yan2018stgcn} utilize a spatial configuration partitioning strategy based on the inherent locality of the human skeleton. The \emph{root node} refers to the joint under consideration for defining the local neighborhood in the graph. For each root node, its neighboring nodes are divided into three subsets:
\begin{enumerate}
    \item The root node itself, representing self-connections.
    \item The centripetal group: neighboring nodes closer to the skeleton's gravity center than the root node.
    \item The centrifugal group: nodes farther from the gravity center compared to the root node.
\end{enumerate}
The gravity center at each frame is computed as the average coordinate of all joints, reflecting the natural concentric and eccentric motions of body parts. Each of the three spatial configurations defined in this partitioning strategy are represented by separate adjacency matrices $\textbf{A}_k$, where $k \in \{1, 2, 3\}$.

\subsection{Spatio-Temporal Graph Convolutional Block}

ST-GCN operates on the graph by applying both spatial and temporal convolutions. The spatial graph convolution is defined as:
\begin{equation}
\textbf{X}_{\mathit{out}} = \sum_{k=1}^{K_v} \textbf{W}_k (\textbf{X}_{\mathit{in}} \textbf{A}_k)
\end{equation}
where $\textbf{X}_{\mathit{in}}$ and $\textbf{X}_{\mathit{out}}$ are the input and output feature tensors, $\textbf{W}_k \in \mathbb{R}^{C_{\mathit{out}} \times C_{\mathit{in}} \times 1 \times 1}$ is a weight matrix for $1 \times 1$ convolution, and $\textbf{A}_k \in \mathbb{R}^{V \times V}$ is the adjacency matrix representing the connections between nodes for the $k$-th spatial configuration. With the partitioning strategy described above, $K_v$ is set to 3.

The adjacency matrix $\textbf{A}_k$ is normalized using the following equation:
\begin{equation}
\textbf{A}_k = \Lambda_k^{-\frac{1}{2}} \bar{\textbf{A}}_k \Lambda_k^{-\frac{1}{2}}
\end{equation}
where $\bar{\textbf{A}}_k \in \mathbb{R}^{N \times N}$ is the unnormalized adjacency matrix, and $\Lambda_k$ is the diagonal degree matrix, with $\Lambda_{ii}$ representing the degree of node $i$. Normalization ensures that feature aggregation is scaled appropriately, preventing high-degree nodes from dominating the output.

A Temporal Convolutional Network (TCN) is then applied to $\textbf{X}_{out}$ to capture the temporal dynamics in the data. Temporal convolutions operate along the time dimension, leveraging the fixed temporal connectivity where each joint is linked to the same joint in adjacent frames. This setup allows the use of a standard $K_t \times 1$ convolution on the $C \times T \times V$ feature maps.

\begin{figure*}[t]
    \centering
    \includegraphics[width=\textwidth]{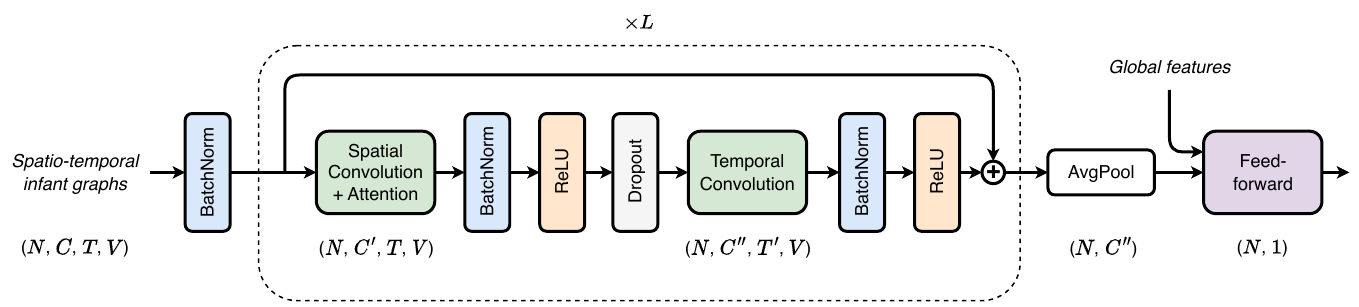}
    \caption{Illustration of the network architecture. The diagram shows the ordering of layers in the STGCN/AAGCN models. The dimensions of the input tensor (including batch size), are highlighted underneath those layers that affect them.}
    \label{fig:network_architecture}
\end{figure*}

\subsection{Adaptive Graph Convolutional Block}

Adaptive Graph Convolutional Networks (AAGCN)~\cite{shi2020aagcn} extend the capabilities of ST-GCN by introducing a learnable adaptive graph structure. The adaptive convolution in AAGCN is expressed as:
\begin{equation}
\textbf{X}_{out} = \sum_{k=1}^{K_v} \textbf{W}_k \textbf{X}_{in} (\textbf{B}_{k} + \alpha \textbf{C}_k)
\end{equation}
where $\textbf{B}_{k} \in \mathbb{R}^{V \times V}$ is the global graph learned from the data, initialized with the human-body-based adjacency matrix and updated during training without constraints. $\textbf{C}_{k} \in \mathbb{R}^{V \times V}$ represents the individual graph adapted to each sample, modulated by a parameterized coefficient $\alpha$ unique to each graph convolution layer in the model.

AAGCN incorporates an STC (Spatial-Temporal-Channel) attention module, where per-sample attention weights are learned to express the importance of joints, time steps, and feature channels. The attention masks are applied sequentially to the input tensor, always in a residual manner~\cite{wang2017residual}.

\emph{Spatial attention} enhances the importance of specific joints by learning a weight mask $\mathbf{M}_s \in \mathbb{R}^{1 \times 1 \times V}$. This mask is computed as:
\begin{equation}
\label{eq:ms}
\mathbf{M}_s = \sigma(\mathbf{W}_s(\text{AvgPool}_T(\mathbf{X}))),
\end{equation}
where $\mathbf{X} \in \mathbb{R}^{C \times T \times V}$ is the input tensor, $\text{AvgPool}_T$ averages over the temporal dimension, $\mathbf{W}_s \in \mathbb{R}^{1 \times C \times K_s}$ are the learnable weights of a 1D convolution, and $\sigma$ is the sigmoid activation function.

\emph{Temporal attention} highlights the importance of specific frames by learning a weight mask $\mathbf{M}_t \in \mathbb{R}^{1 \times T \times 1}$. The mask is computed as:
\begin{equation}
\mathbf{M}_t = \sigma(\mathbf{W}_t(\text{AvgPool}_V(\mathbf{X}))),
\end{equation}
where $\text{AvgPool}_V$ averages over the spatial dimension and the definitions of the other symbols are similar to Eq.~\eqref{eq:ms}.

\emph{Channel attention} assigns importance to feature channels by learning a weight vector $\mathbf{M}_c \in \mathbb{R}^{C \times 1 \times 1}$. This mechanism employs a squeeze-and-excitation operation~\cite{hu2018se}, which first compresses global information across temporal and spatial dimensions and then recalibrates the channel-wise features. The attention weights are computed as:
\begin{equation}
\mathbf{M}_c = \sigma(\mathbf{W}_2(\delta(\mathbf{W}_1( \text{AvgPool}_{T,V}(\mathbf{X}))))),
\end{equation}
where $\text{AvgPool}_{T,V}$ averages over both temporal and spatial dimensions, $\mathbf{W}_1 \in \mathbb{R}^{\frac{C}{r} \times C}$ and $\mathbf{W}_2 \in \mathbb{R}^{C \times \frac{C}{r}}$ are the weights of two fully connected layers, $\delta$ denotes the rectified linear unit (ReLU) activation function, and $r=2$ is the reduction ratio.

Each AAGCN block comprises the spatial GCN with the STC-attention module, followed by a TCN module identical to that in the STGCN.

\subsection{Multi-Stream Model}

Multi-Stream AAGCN (MS-AAGCN)~\cite{shi2020aagcn} leverages multiple feature streams to enhance prediction accuracy. Unlike the original work, where each stream is input into separate AAGCN models and their classification scores are fused using a weighted sum, our approach encodes multiple streams as node features within a single infant graph. The features we integrate include joint coordinates, bone directions, joint velocities, and joint accelerations. Bone direction is calculated as the vector difference between connected joints in the graph, while velocity and acceleration are derived from changes in joint positions across consecutive frames, capturing the dynamical aspects of the motor sequence. Consequently, the channel dimension in MS-AAGCN is the product of the number of streams and the number of spatial coordinates considered.

\subsection{Network Architecture}

The complete network architecture is illustrated in Figure~\ref{fig:network_architecture}. Both the spatial GCN components (including the STC module for AAGCN) and the TCN components are followed by a batch normalization (BatchNorm)~\cite{ioffe2015batchnorm} layer and a ReLU activation layer. To enhance training stability and ensure efficient gradient flow, each block incorporates a residual connection. Additionally, dropout with a rate of 0.5 is applied between the GCN and TCN layers to reduce overfitting. The network starts with a data BatchNorm layer to standardize the input.

The architecture consists of a total of 3 STGCN or AAGCN blocks, each with latent dimension 256. The temporal stride for the TCN is set to 1, maintaining sequence length across layers. Following the stack of spatio-temporal convolution blocks, average pooling is applied to reduce the dimensionality of the feature maps and aggregate spatial information.

After pooling, global features derived from the infant recordings, as discussed in Section~\ref{sec:ml_baseline}, may be concatenated with the pooled graph features. The combined features are then fed through a two-layer feedforward network with hidden dimensions of 64 and 32, employing a ReLU activation function in between. The final linear layer outputs dimensionality 1 in order to make global predictions for the entire input graph.

\subsection{Training and Implementation}

The training goal is to minimize the discrepancy between the model's predictions and the actual ages of the infants. The optimization objective is defined by the Mean Squared Error (MSE) loss function:

\begin{equation}
\mathcal{L} = \frac{1}{N} \sum_{i=1}^{N} (\age_i - \hatage_i)^2,
\end{equation}

where $\age_i$ denotes the true age of the infant, $\hatage_i$ is the predicted age by the model, and $N$ is the number of samples in each batch. We employ the MoMo optimizer~\cite{schaipp2023momo} with a learning rate (lr) of 0.01 to minimize the loss. MoMo provides an adaptive lr strategy compatible with any momentum method, reducing the need for extensive lr tuning.

For model evaluation, we utilize $k$-fold cross-validation with $k=10$, ensuring the data is split according to subjects to obtain a robust performance assessment. During each fold, models are trained for 20 epochs with a batch size of 32, and the best-performing checkpoint on the validation set is saved. The models are implemented using the PyTorch Lightning framework~\cite{falcon2019lightning}, which facilitates efficient training, logging, and evaluation with minimal boilerplate code. All training and inference tasks were conducted on an Nvidia V100 GPU with 32GB of memory, utilizing 8 data loader workers running on Intel Xeon Gold 6230 CPUs.

To make predictions, we employ the same bagging approach as before by averaging the outputs from the 10 models trained through 10-fold cross-validation. Running all 10 models on the set of 14 severe infants, divided into 151 segments, each 600 timesteps long, on an Nvidia V100 GPU, takes 48 seconds using 2D data and 57 seconds with 3D data, roughly equivalent to 3 seconds per 2D sample and 4 seconds per 3D sample.

\section{Machine Learning Baseline}
\label{sec:ml_baseline}

To establish a baseline for our deep learning models, we utilized ten computational indexes previously developed to quantify infant limb movements~\cite{marchi2020movement}. These indexes capture movement characteristics across three main categories: coordination, movement distance, and overall movement quality.

\paragraph{Coordination Indexes}
We adopt the methodology proposed by Kanemaru et al.~\cite{kanemaru2012increasing} to measure limb coordination. This involves analyzing coordination between the right and left hands, as well as between the right and left feet. The first canonical correlations between these limb positions are calculated, along with the respective angles between the first canonical vectors. Additionally, the three-dimensional tangential velocities for both hands and feet are computed. Pearson's correlation coefficients between these velocities are used to quantify the degree of synchronization in limb movements. The analysis focuses on two types of correlations: limb position correlations and limb velocity correlations. Fisher's Z score is used to compare these correlation coefficients, offering a standardized measure for assessing coordination in relation to age and limb combinations.

\paragraph{Distance Indexes}
To assess how closely infants can bring their limbs together, we calculate distances between the right and left hand markers, as well as between the right and left foot markers~\cite{miyagishima2016characteristics}. We also evaluate the height displacement of limbs from the surface to measure the ability to raise and maintain hands or feet against gravity. The minimal and maximal distances, represented by the 5th and 95th percentiles, are used to capture the range of limb movements. Additionally, we compute the probabilities of limbs being in close proximity (less than 0.25m apart) and of limbs being lifted above a 0.1m height threshold. These probabilities, referred to as close-limb and lift-limb probabilities, provide insights into the frequency and extent of limb proximity and antigravity postures.

\paragraph{Global Movement Quality Indexes}
Hjorth parameters—activity, mobility, and complexity—are employed as non-linear measures for the time-series analysis of limb movements~\cite{hjorth1970eeg}. These parameters are calculated from velocity, acceleration, and jerk data, respectively, using a sliding window approach with a one-second duration and a 50\% overlap. To minimize transient effects, the first and last windows are discarded. The parameters are averaged across windows, spatial components, and body sides to quantify movement dynamics for both hands and feet. The activity parameter estimates the variance in velocity, mobility measures the standard deviation of acceleration relative to velocity, and complexity represents the change in frequency by comparing the signal's similarity to a pure sine wave.

To account for the variability in segment lengths, a weighted average is applied to each subject's features in the dataset. Each feature is multiplied by a weight, which corresponds to the relative length of the annotated segment it belongs to, compared to the total length of all segments for a given subject. This approach ensures that segments with more data have a greater influence on the overall feature representation of each subject.

Table~\ref{tab:correlation_results} presents the correlation values for the motor indices examined in this study. The Spearman correlation method is used instead of the standard Pearson correlation due to the non-linear, stepwise nature of developmental skills, such as the ability to bring hands together. These skills often develop in a monotonic manner, making Spearman correlation more suitable for analysis. The expansion of the dataset to include a larger number of recorded subjects compared to the earlier study~\cite{marchi2020movement} has resulted in lower correlation scores. This decrease is likely due to the broader population studied here, which exhibits greater variability in movement patterns compared to the curated selection of recordings used previously.

\begin{table}[ht]
\caption{List of features used for prediction of corrected postmenstrual age, divided into indexes of coordination, indexes of distance and indiexes of global movement quality. The Spearman correlation $\rho$ measure the strength and direction of the monotonic relationship between each parameter and the corrected postmenstrual age.}
\label{tab:correlation_results}
\centering
\begin{tabular}{lcc}
\toprule
Feature & $\rho_{\text{hands, age}}$ & $\rho_{\text{feet, age}}$ \\
\midrule
Position canonical correlation (Z-score) & 0.18 & 0.16 \\
Angles between canonical vectors & -0.29 & -0.23 \\
Velocity correlation (Z-score) & 0.13 & 0.12 \\
\midrule
Distance (< 5th percentile) & 0.01 & -0.19 \\
Lift (> 95th percentile) & 0.17 & 0.06 \\
Close-limb probability & -0.09 & -0.01 \\
Lift-limb probability & 0.16 & 0.07 \\
\midrule
Hjorth activity & -0.08 & -0.15 \\
Hjorth mobility & -0.06 & 0.13 \\
Hjorth complexity & -0.19 & 0.06 \\
\bottomrule
\end{tabular}
\end{table}

We use four common machine learning models to predict the corrected age: 1. Linear Regression (LR), 2. Random Forest (RF)~\cite{breiman2001randomforest}, 3. Gradient Boosting (GB)~\cite{friedman2001gradientboosting}, and 3. XGBoost (XGB)~\cite{chen2016xgboost}. Using only hands complexity and lift-hands probability as inputs to LR as in \cite{marchi2020movement} faired worse than using all available features, so we continued the analysis with the full dataset. We again use 10-fold cross validation with the exact same data splits as in Section~\ref{sec:method} to train the models. In addition to this, hyperparameter tuning is performed for the three latter models: RF, GB and XGB. Grid search is implemented such that we search for the optimal configuration for minimizing the MSE loss across the folds. The search space is defined as follows:
\begin{itemize}
     \item Number of trees (RF) / boosting stages (GB) / gradient boosted trees (XGB): $n \in \{100, 300, 500, 700, 900\}$,
    \item Maximum depth of trees: $d \in \{1, 3, 5, 7, 9\}$,
    \item Learning rate (GB, XGB): $lr \in \{0.1, 0.15, 0.2, 0.25\}$.
\end{itemize}
The best-performing configurations of each model were identified as RF with $n = 300$, $d = 7$, GB with $n = 300$, $d = 1$, $lr = 0.2$, and XGB with $n = 700$, $d = 1$, $lr = 0.15$.

\section{Experiments}

\subsection{Temporal Kernel Size}

\begin{figure*}[ht]
    \centering
    \includegraphics[width=\textwidth]{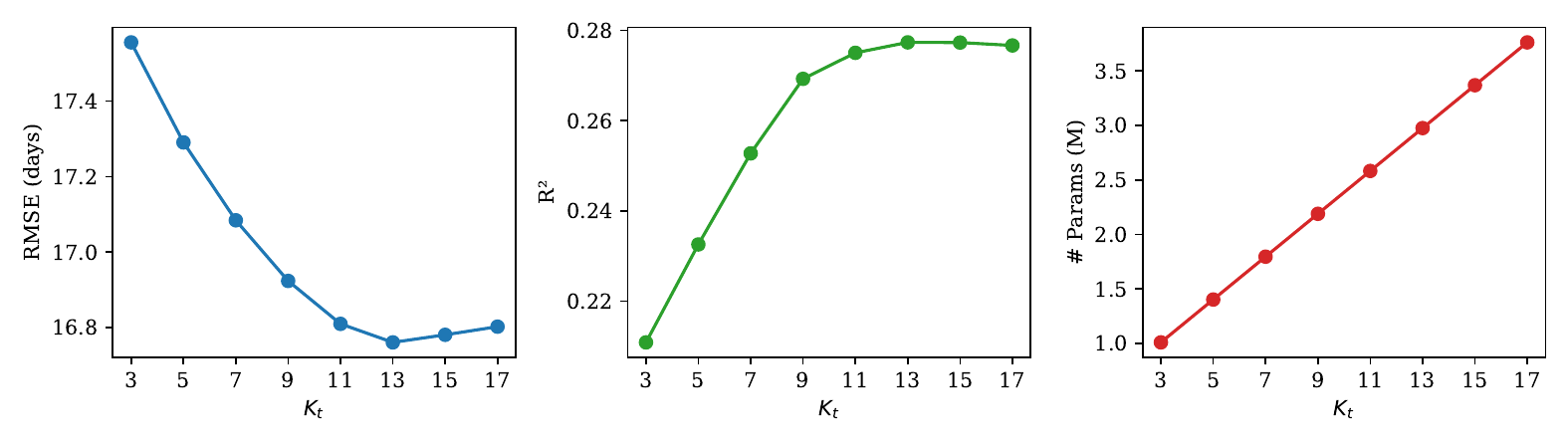}
    \caption{Evaluation of the STGCN model performance as a function of temporal kernel size. The left plot shows the RMSE against $K_t$ the middle plot depicts $R^2$ scores, and the right plot illustrates how the number of model parameters increases linearly for larger convolution kernels.}
    \label{fig:kt}
\end{figure*}

To determine the optimal temporal kernel size ($K_t$) for the STGCN, we conducted a systematic scan of $K_t$ values ranging from 3 to 17, evaluating every odd integer in that interval. As shown in Figure~\ref{fig:kt}, we observe that the Root Mean Square Error (RMSE) decreases rapidly up to $K_t = 13$ and then stagnate for larger kernel sizes. The $R^2$ score also reaches its optimal value at $K_t = 13$.

This behavior may have to do with a trade-off between temporal context and model complexity. Smaller $K_t$ values capture only short-term temporal dependencies, which may not fully describe the infant movements. Increasing the temporal kernel size too much could lead to excessive temporal smoothing or overfitting, where the model captures noise or irrelevant long-range dependencies. 

In addition to evaluating RMSE and $R^2$, we analyzed the model's parameter count as a function of $K_t$. As shown in Figure~\ref{fig:kt}, the number of parameters grows linearly with $K_t$. The parameter count for a 1D convolutional layer is given by $\text{Params} = (C_{\text{in}} \times K_t + 1) \times C_{\text{out}}$, where $C_{\text{in}}$ and $C_{\text{out}}$ represent the number of input and output channels, respectively, and $K_t$ is the kernel size. The $+1$ term accounts for the bias added to each filter. As $K_t$ increases, the number of parameters rises proportionally, assuming $C_{\text{in}}$ and $C_{\text{out}}$ are constant. In our case, the complete STGCN model has approximately 1 M parameters for $K_t = 3$, and surpasses 3.5 M for $K_t = 17$.

The selected temporal kernel size is higher than the more conventional $K_t \in \{7, 9\}$, found as optimal values for action classification~\cite{cheng2020skeleton, zhang2020context}. This difference may arise from the slower, more gradual nature of infant movements compared to the faster and more discrete actions in the datasets used for action classification. Given these findings, we use $K_t = 13$ in all subsequent experiments.

\subsection{Graph Adaptivity}

In this experiment, we aim to quantify the benefits of adaptive graph structures in improving the performance of infant movement analysis. To do so, we conduct an ablation study comparing three models: STGCN (static adjacency matrix), AGCN (with a fully adaptable adjacency matrix), and AAGCN (with an adaptive graph and attention mechanisms active).

As summarized in Table~\ref{tab:adaptivity}, AGCN shows an improved RMSE and $R^2$ over STGCN by learning a fully adaptive adjacency matrix, which allows the model to establish new connections that are not constrained by the predefined structure of the human body.

\begin{table}[h]
\centering
\caption{Performance comparison for increasingly adaptive graphs}
\label{tab:adaptivity}
\begin{tabular}{lccc}
\toprule
Model & \# Params & RMSE & $R^2$ \\
\midrule
STGCN & 2.976 M & 16.76 $\pm$ 2.53 & 0.28 $\pm$ 0.18 \\
AGCN & 3.178 M & 16.67 $\pm$ 3.09 & 0.29 $\pm$ 0.18 \\
AAGCN & 3.395 M & 16.46 $\pm$ 2.53 & 0.30 $\pm$ 0.19 \\
\bottomrule
\end{tabular}
\end{table}

The introduction of the STC attention module in AAGCN further enhances performance, demonstrating the importance of dynamically weighting different parts of the graph during training. The STC mechanism allows AAGCN to adapt to each subject’s unique movements by learning which nodes, temporal steps, and channels are most relevant for age prediction. AAGCN achieves the lowest error rates in this experiment, and is the chosen model for the rest of the study.

\subsection{Graph Initialization}

In addition to varying the input streams, we explored how different graph initialization strategies affect the AAGCN model performance. Specifically, we tested three initialization schemes: physical (based on the human skeletal structure), coordination (which adds six bidirectional edges between the hands and feet), and fully connected (where all nodes are connected). The normalized version of the centrifugal group in the graph partitioning of these initializations are shown in Figure~\ref{fig:init}. The normalization assign weights according to the number of incoming edges to a node as to not over emphasize e.g. the neck joint with many incoming edges when sum pooling is used in the GCN. The fully connected graphs has notably low uniform edge weights after normalization as each row in the adjacency matrix should add up to one.

\begin{table}[h]
\centering
\caption{Performance comparison for three different AAGCN graph initialization strategies}
\label{tab:graph}
\begin{tabular}{lccc}
\toprule
Initialization & \# Params & RMSE (days) & $R^2$ \\
\midrule
Physical & 3.395 M & 16.46 $\pm$ 2.53 & 0.30 $\pm$ 0.19 \\
Coordination & 3.395 M& 16.39 $\pm$ 2.94 & 0.30 $\pm$ 0.19 \\
Fully-connected & 3.395 M & 16.66 $\pm$ 2.23 & 0.29 $\pm$ 0.15 \\
\bottomrule
\end{tabular}
\end{table}

As shown in Table~\ref{tab:graph}, both the physical and coordination initializations outperform the fully connected graph initialization. The coordination graph initialization achieves a slightly better average RMSE across the folds compared to the physical initialization, while both yield identical $R^2$ scores to two decimal places. These results suggest that the predictive power when using the physical and coordination graphs is quite similar. However, given the marginally lower RMSE, we chose the coordination graph initialization for all further experiments.

The motivation for trying the coordination-based initialization stems from the role of limb coordination in normal development. Upper and lower limbs must work together for key movements, such as crossing the midline or touching, as well as for complex full-body actions like crawling and walking. This experiment demonstrates that incorporating an inductive bias reflecting the human skeletal structure, and also limb coordination, can be beneficial for modeling infant movement. In contrast, the fully connected graph, which lacks inductive biases, performs the worst. The lack of prior structure in this approach may make it difficult for the model to learn meaningful relationships between the joints, as seen in Figure~S1. However, caution should be exercised when trying to extract explainability from such visualizations, as the highly dimensional latent space obscures any direct interpretations.

\subsection{Input Streams}

We further evaluate how different input streams impact the performance of MS-AAGCN by progressively incorporating features beyond joint positions (J). These additional streams include bone directional vectors (B), velocity (V), and acceleration (A) of joints. As Table~\ref{tab:streams} illustrates, the addition of bone directional vectors (JB-AAGCN) yields the most significant performance improvement, achieving the lowest RMSE. This result is in line with previous work finding joint and bone direction important input features for action recognition~\cite{shi2019twostream}, and histograms of joint directions has also been used as input for classifying abnormal infant movements~\cite{mccay2020binaryab}.

\begin{table}[h]
\centering
\caption{Performance comparison of input stream configurations}
\label{tab:streams}
\begin{tabular}{lccc}
\toprule
Model & \# Params  & RMSE (days) & $R^2$ \\
\midrule
J-AAGCN & 3.395 M & 16.39 $\pm$ 2.94 & 0.30 $\pm$ 0.19 \\
JB-AAGCN & 3.400 M & 16.18 $\pm$ 2.94 & 0.32 $\pm$ 0.20 \\
JBV-AAGCN & 3.404 M & 16.31 $\pm$ 2.72 & 0.31 $\pm$ 0.20 \\
JBVA-AAGCN & 3.408 M & 16.75 $\pm$ 3.14 & 0.27 $\pm$ 0.21 \\
\bottomrule
\end{tabular}
\end{table}

The subsequent addition of velocity (JBV-AAGCN) and acceleration (JBVA-AAGCN) streams yield slightly worse results. The increased error especially for JBVA-AAGCN could be attributed to the challenge of modeling higher-order derivatives of position, which may amplify artifacts or irrelevant fluctuations in the data. Based on this, we choose JB-AAGCN as our final model for age assessment.

\subsection{Data Dimensionality and Rotation}

JB-AAGCN 2D was trained on data preprocessed to include only the $(x, y)$-coordinates of joint positions, reflecting the common 2D filming approach used for infant age estimation. In contrast, JB-AAGCN 3D was trained with all three $(x, y, z)$-coordinates available from the 3D filming setup. The 3D model achieved lower error rates across all metrics compared to the 2D version, demonstrating the benefits of using 3D data for more accurate age estimation.

\begin{table}[h]
\centering
\caption{Performance comparison of data dimensionality and rotation ablation for the JB-AAGCN model.}
\label{tab:rot}
\begin{tabular}{lccc}
\toprule
Data & \# Params & RMSE (days) & $R^2$ \\
\midrule
2D rotated & 3.398 M & 16.96 $\pm$ 2.81 & 0.27 $\pm$ 0.18 \\
2D w/o rot. & 3.398 M & 17.32 $\pm$ 3.34 & 0.23 $\pm$ 0.21 \\
3D rotated & 3.400 M & 16.39 $\pm$ 2.94 & 0.30 $\pm$ 0.19 \\
3D w/o rot. & 3.400 M & 16.51 $\pm$ 3.68 & 0.29 $\pm$ 0.23 \\
\bottomrule
\end{tabular}
\end{table}

Additionally, we conduct an ablation study to assess the effect of removing rotation during preprocessing. Both the 2D and 3D models experienced performance degradation when rotation was excluded, underscoring the importance of aligning the poses in order to mitigate discrepancies caused by varying filming angles. We use rotated data when analyzing the models clinical impact in the next section.

\section{Results}

The results of the 10-fold cross-validation are shown in Table~\ref{tab:k_fold_cv_results}. Here we report the average validation metrics across the folds, including Mean Error (ME), Root Mean Squared Error (RMSE), Mean Absolute Error (MAE), $R^2$, and Pearson correlation, along with their corresponding standard deviations.

\begin{table*}[ht]
\centering
\caption{Summary of model performance from 10-fold cross-validation.}
\label{tab:k_fold_cv_results}
\begin{tabular}{lccccc}
\toprule
Model & ME (days) & RMSE (days) & MAE (days) & $R^2$ & Pearson \\
\midrule
Linear Regression & -0.11 $\pm$ 5.52 & 19.50 $\pm$ 2.49 & 14.94 $\pm$ 1.61 & 0.03 $\pm$ 0.23 & 0.39 $\pm$ 0.22 \\
Random Forest & -0.42 $\pm$ 4.58 & 18.86 $\pm$ 3.39 & 13.98 $\pm$ 2.57 & 0.10 $\pm$ 0.23 & 0.41 $\pm$ 0.27 \\
Gradient Boosting & -0.15 $\pm$ 4.06 & 18.12 $\pm$ 3.56 & 14.07 $\pm$ 2.85 & 0.15 $\pm$ 0.26 & 0.48 $\pm$ 0.20 \\
XGBoost & -0.07 $\pm$ 3.59 & 18.02 $\pm$ 3.73 & 14.30 $\pm$ 2.57 & 0.16 $\pm$ 0.27 & 0.49 $\pm$ 0.20 \\
\midrule
JB-AAGCN 2D & 0.22 $\pm$ 3.73 & 16.96 $\pm$ 2.81 & 12.49 $\pm$ 2.36 & 0.27 $\pm$ 0.18 & 0.57 $\pm$ 0.16 \\
JB-AAGCN 3D & -0.54 $\pm$ 3.20 & 16.18 $\pm$ 2.94 & 11.78 $\pm$ 1.98 & 0.32 $\pm$ 0.20 & 0.61 $\pm$ 0.20 \\
JB-AAGCN 3D + features & -1.08 $\pm$ 3.46 & 16.32 $\pm$ 2.30 & 12.69 $\pm$ 2.02 & 0.32 $\pm$ 0.14 & 0.60 $\pm$ 0.11 \\
\bottomrule
\end{tabular}
\end{table*}

Among the machine learning baseline models, the ensemble-based methods clearly outperformed the linear regression model, with XGBoost showing the best overall performance. However, the ensemble models exhibit greater variance across the folds compared to linear regression, suggesting that they may be more sensitive to variations in the data splits. Deep learning models, shown in the lower section of Table~\ref{tab:k_fold_cv_results}, consistently outperformed the machine learning baselines. Among the deep learning approaches, JB-AAGCN 3D achieved the lowest RMSE at 16.18 days and MAE at 11.78 days, as well as the highest Pearson correlation of 0.61. We can also note that a purely data driven approach performs better than adding manually engineered motion features to the feedforward head of the model, denoted as JB-AAGCN 3D + features in Table~\ref{tab:k_fold_cv_results}.

The prediction of age becomes particularly challenging for the youngest infants, as shown by the scatter plots in Figure~S2 and Figure~S3. Infants younger than 70 days exhibit higher prediction errors across all models. Younger infants' exhibit more time-to-time and interindividually variable movements, while their age-related change is less apparent, making an accurate algorithmic age prediction more difficult, if not impossible. As a result, models struggle to generalize well for this group, and their predictive accuracy improves as the infants get older and their movement patterns become more stable and coordinated.

\subsection{Results by Location}
Generalization of the model performance was indirectly assessed by comparing the prediction errors between study sites (Table~\ref{tab:location_comparison_all_ages} and Table~\ref{tab:location_comparison_older_than_80}). With the full cohorts from both sites, the prediction error was clearly larger in the Pisa data. However, the apparent site difference was due to a higher proportion of younger infants in the  Pisa data, and the findings were essentially comparable between sites for infants over 80 days.  Notably, the prediction error was systematically higher in the 2D models compared to the 3D models in both sites. 

\begin{table}[h]
\centering
\caption{Performance for the JB-AAGCN model across locations for infants of all ages. The sample size is 97 for the Pisa cohort, and 62 for the Helsinki cohort.}
\label{tab:location_comparison_all_ages}
\begin{tabular}{ccccc}
\toprule
Data & Location & Corrected Age & RMSE & Pearson \\
\midrule
\multirow{2}{*}{2D} & Pisa & 85.86 $\pm$ 22.23 & 18.61 $\pm$ 3.57 & 0.52 $\pm$ 0.18 \\
                    & Helsinki & 100.32 $\pm$ 15.31 & 13.20 $\pm$ 3.29 & 0.18 $\pm$ 0.54 \\
\midrule
\multirow{2}{*}{3D} & Pisa & 85.86 $\pm$ 22.23 & 17.89 $\pm$ 4.73 & 0.54 $\pm$ 0.27 \\
                    & Helsinki & 100.32 $\pm$ 15.31 & 10.86 $\pm$ 3.11 & 0.51 $\pm$ 0.40 \\
\bottomrule
\end{tabular}
\end{table}

\begin{table}[h]
\centering
\caption{Performance for the JB-AAGCN model across locations for infants older than 80 days. The sample size is 74 for the Pisa cohort, and 59 for the Helsinki cohort.}
\label{tab:location_comparison_older_than_80}
\begin{tabular}{ccccc}
\toprule
Data & Location & Corrected Age & RMSE & Pearson \\
\midrule
\multirow{2}{*}{2D} & Pisa & 96.18 $\pm$ 10.44 & 11.57 $\pm$ 4.09 & 0.27 $\pm$ 0.46 \\
                    & Helsinki & 101.42 $\pm$ 14.86 & 13.22 $\pm$ 3.88 & 0.20 $\pm$ 0.56 \\
\midrule
\multirow{2}{*}{3D} & Pisa & 96.18 $\pm$ 10.44 & 10.85 $\pm$ 4.27 & 0.29 $\pm$ 0.31 \\
                    & Helsinki & 101.42 $\pm$ 14.86 & 10.83 $\pm$ 3.17 & 0.52 $\pm$ 0.41 \\
\bottomrule
\end{tabular}
\end{table}

\subsection{Clinical Proof of Concept}

\begin{figure*}[ht]
    \centering
    \includegraphics[width=.97\textwidth]{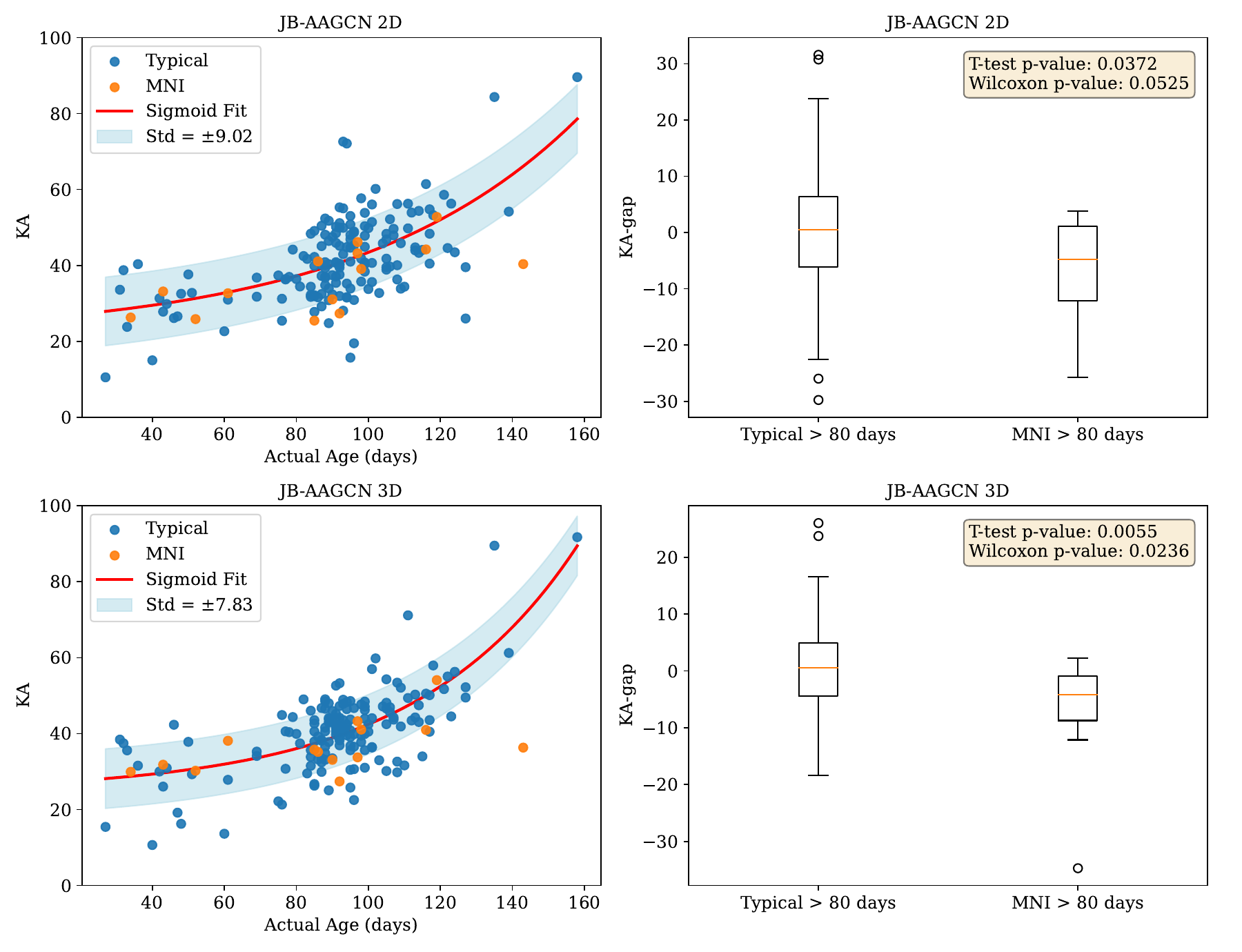}
    \caption{Left: KA vs actual corrected age, with a fitted sigmoid curve in red. Blue markers represent typically developing infants, while orange markers indicate infants with MNI. Right: Boxplots for the residuals (difference to the sigmoid curve) for both groups, focusing on infants older than 80 days. The results show a separation between infants with a typical development vs MNI, with the latter showing a lower KA.}
    \label{fig:predictions}
\end{figure*}

To demonstrate the clinical applicability of our approach, we used the two best-performing models, JB-AAGCN 2D and JB-AAGCN 3D, to predict the age. The predictions were converted into a unitless index by normalizing them to the postnatal age range of 50 to 150 days. This normalization enabled the generation of \emph{KA growth charts}, which in typically developing infants showed an initially flat trajectory followed by a monotonic rise (see the two sigmoid fits in Figure~\ref{fig:predictions}). Additionally, we computed KA for a group of $N = 14$ infants who were later diagnosed with MNI.

The overall findings in the infants with MNI were comparable between 3D and 2D models: The KA estimates in the younger infants were within the levels seen in the typically developing infants. However, the older infants with MNI did not show the age-dependent KA increase that was clear in the typically developing infants. A straightforward individual level measure of normality could be the KA-gap, defined as the difference between infant's KA and the age-related mean KA in the typically developing infants (red line in the KA growth charts). We found a statistically significant difference in the KA-gap between typically developing infants and those with MNI as seen from the two boxplots in Figure~\ref{fig:predictions}. The observations together suggest that KA, or its age-adjusted estimate KA-gap, may differentiate between typically and atypically developing infants.

\section{Discussion}

Our results suggest a generally better performance with 3D than 2D data. However, the difference is perhaps smaller than intuitively expected. It is important to note that in the present context our 2D dataset was obtained from the 3D pose estimation by removing the depth only. The accuracy of KA estimates in 2D data may be more compromised if the pose estimation was already based on 2D video only~\cite{groos2022development}. This needs to be tested on prospective datasets. 

The developmental change in the KA index, i.e. the KA growth charts, observed in our limited population suggests a monotonic but non-linear development of an infants' kinetics. This observation aligns well with the everyday clinical experience that the very young infants express widely ranging motor activity patterns where specific components, such as fidgety movements~\cite{einspieler2005prechtl}, may reflect underlying neurological abnormalities without specific links to their actual chronological age.  

Our observations from the limited cohort of infants with neurodevelopmental problems suggest that developmental trajectory of the overall kinetics may be initially age-typical, and beginning to deviate only during later months. This is fully in line with the idea that the early spontaneous movements, including the GMA-related fidgety movements near three months of age, are generated by the essentially hard-wired pattern generator systems~\cite{einspieler2005prechtl}. Thereafter, the more complex brain mechanisms with cortical and basal ganglia contributions will gradually rate more elaborated movement skills that characterize the later gross motor development~\cite{einspieler2005prechtl}. Hence, GMA-specific movements can provide an evidence-based, indirect marker of early brain function that predicts later abnormal outcomes without assessing the concurrent maturation per se. In contrast, the gradually maturing motor abilities, the infant kinetics as measured by KA, could directly represent the early neurodevelopmental outcome itself. More longitudinal studies with wider age range are needed to test this rationale, and to examine the utility of KA in the clinical research or care. 

\section{Conclusion}

In this study, we introduced a data-driven approach utilizing pose-estimated infant recordings to predict the kinetic age, highlighting the potential advantages of graph neural networks and the increased efficacy of 3D compared to 2D poses in assessing infants' kinetics. We show that data-driven methods can extract the relevant features and outperforms the conventional methods using heuristic limb movement parameters. Expanding the 3D screenings of infant general movements to an even greater scale would be immensely useful for future endeavors as deep learning is known to benefit from more data to enhance performance.

\section*{Code and Data Availability}

The implementation of the proposed methods is available at \url{https://github.com/deinal/infant-aagcn}, and the associated dataset is stored at: \url{https://doi.org/10.5281/zenodo.14269866}.

\bibliographystyle{ieeetran}
\bibliography{references}

\end{document}


\begin{center}
    \Large \textbf{Supplementary Materials}\\[4em]
    \huge Learning Developmental Age from 3D Infant Kinetics Using Adaptive Graph Neural Networks\\[1em]
    \normalsize Daniel Holmberg, Manu Airaksinen, Viviana Marchi, Andrea Guzzetta,\\Anna Kivi, Leena Haataja, Sampsa Vanhatalo, and Teemu Roos
\end{center}

\newpage
\section*{Supplementary Figure 1}
\begin{figure}[h!]
\centering
\includegraphics[width=.9\textwidth]{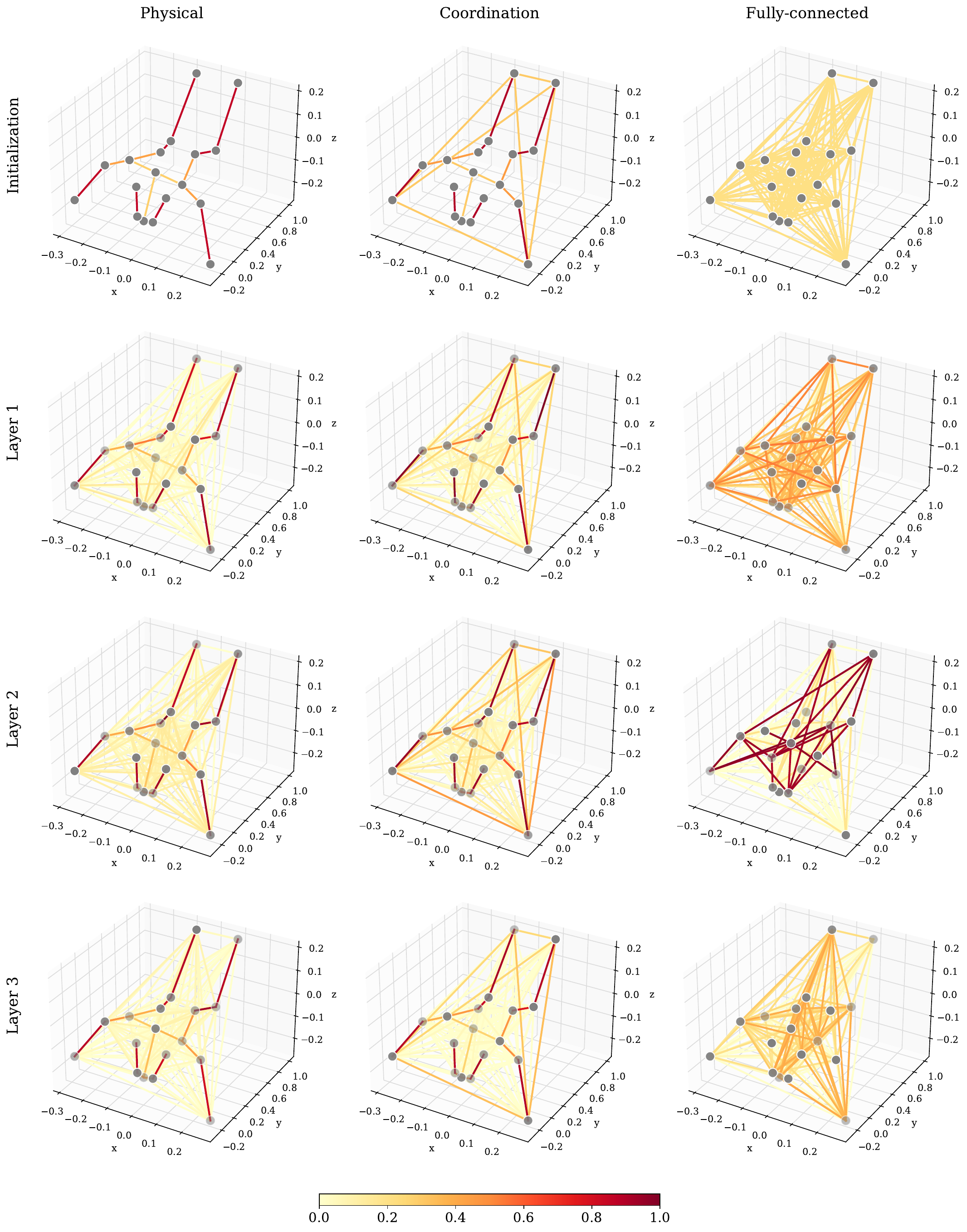}
\caption{Learned connection strengths for different graph initializations in the AAGCN model's centrifugal group of the graph partitioning. Edge weights are normalized between 0 and 1. Node opacity represents the strength of self-connections, scaled between 0.5 and 1.}
\label{fig:learned_graphs_2}
\end{figure}

\newpage
\section*{Supplementary Figure 2}
\begin{figure}[h!]
    \centering
    \includegraphics[width=\textwidth]{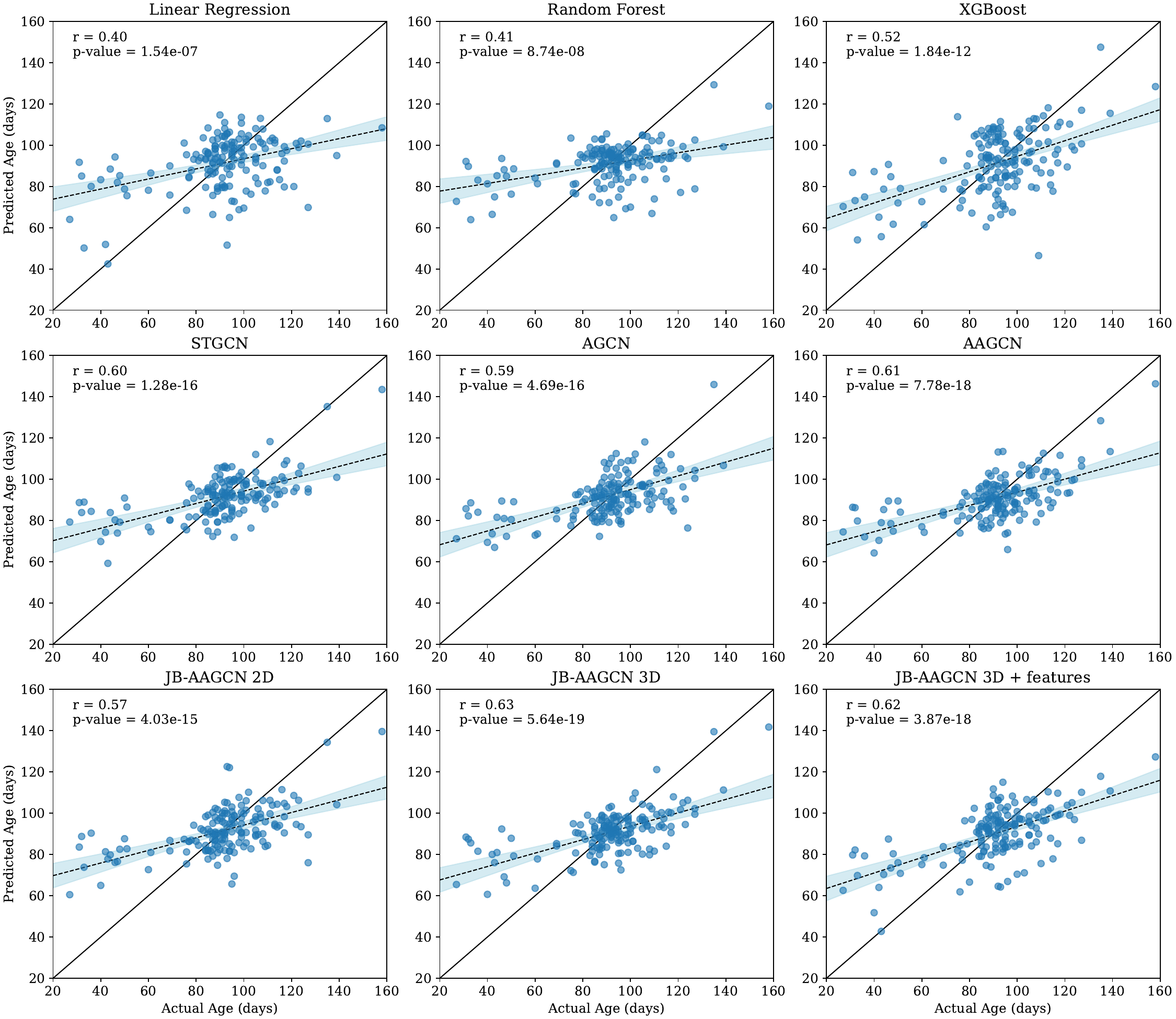}
    \caption{Scatter comparison of predicted vs actual age for model predictions. Each point represents an individual sample's actual corrected age plotted against the predicted age. The black dashed line is a least squares line fit with a 95\% confidence interval in light blue. The solid black line represents the identity line, where predictions would perfectly match actual values. Pearson correlation coefficient (\textit{r}) and p-value are displayed in the upper left corner, indicating the strength of the relationship between predicted and actual ages.}
    \label{fig:comparison}
\end{figure}

\newpage
\section*{Supplementary Figure 3}
\begin{figure}[h]
    \centering
    \includegraphics[width=\textwidth]{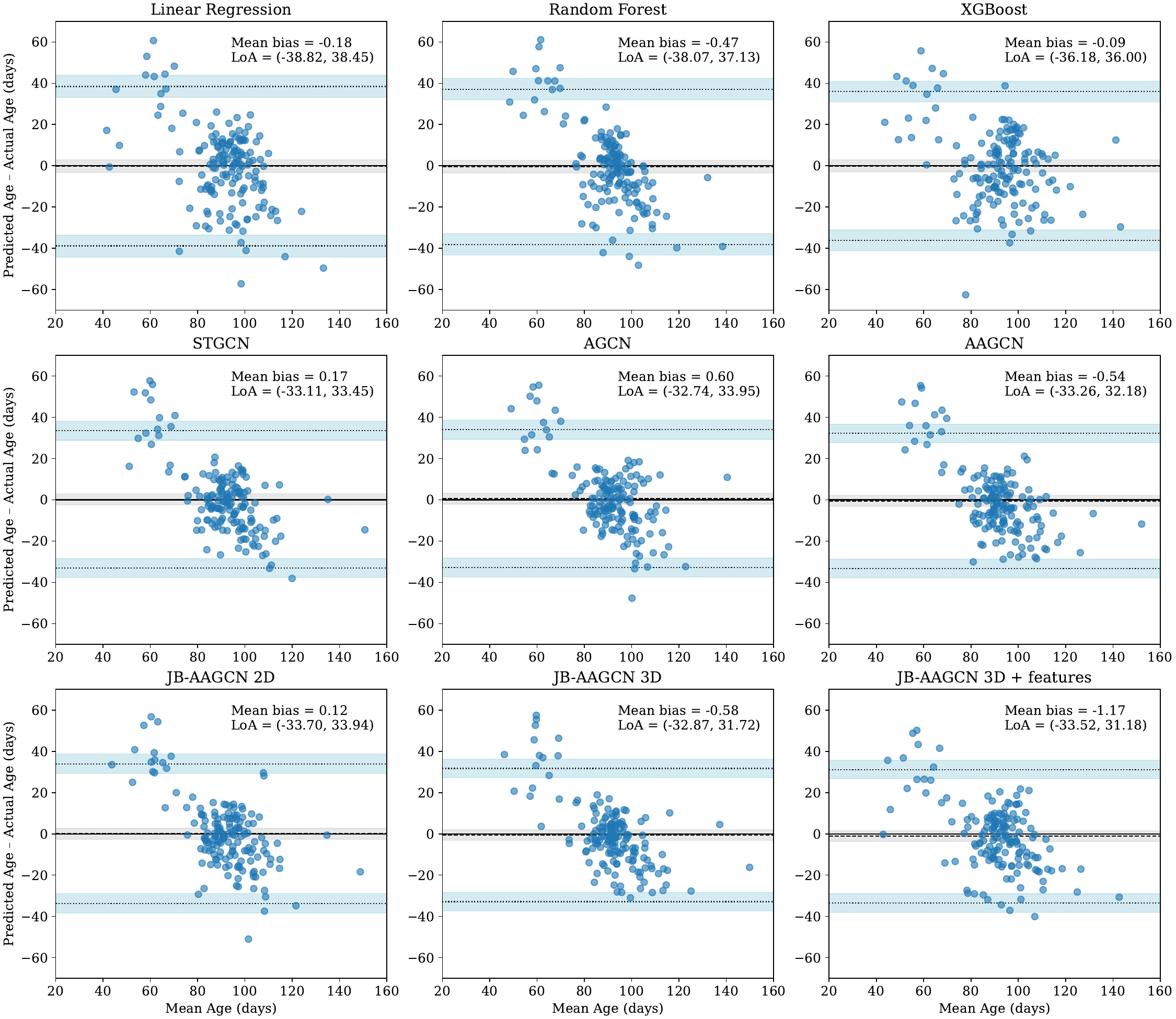}
    \caption{Bland-Altman plots for all models comparing the predicted and actual values. The mean bias (predicted - actual) is shown as a black dashed line, and the limits of agreement (LoA) are calculated as the mean difference $\pm$1.96 times the standard deviation of the differences, shown as black dotted lines. The gray shaded region represents the 95\% confidence interval for the mean bias, while the light blue shaded regions show the 95\% confidence intervals for the LoA.}
    \label{fig:bland_altman}
\end{figure}